\documentclass[runningheads]{llncs}
\usepackage{graphicx}
\usepackage{amsmath}
\usepackage{csvsimple}
\usepackage{algorithm}
\usepackage{algorithmic}

\begin{document}
\title{Improving Time Series Classification with Representation Soft Label Smoothing}
\titlerunning{Representation Soft Label Smoothing}
\author{Hengyi Ma\inst{1}\and
Weitong Chen\inst{1} }
\authorrunning{Hengyi Ma et al.}
\institute{\textsuperscript{1}The University of Adelaide\\ 
\email{hengyi.ma@student.adelaide.edu.au}\\
\email{t.chen@adelaide.edu.au}\\
}
\maketitle              

\begin{abstract}

Previous research has indicated that deep neural network based models for time series classification (TSC) tasks are prone to overfitting. This issue can be mitigated by employing strategies that prevent the model from becoming overly confident in its predictions, such as label smoothing and confidence penalty. Building upon the concept of label smoothing, we propose a novel approach to generate more reliable soft labels, which we refer to as representation soft label smoothing. We apply label smoothing, confidence penalty, and our method representation soft label smoothing to several TSC models and compare their performance with baseline method which only uses hard labels for training. Our results demonstrate that the use of these enhancement techniques yields competitive results compared to the baseline method. Importantly, our method demonstrates strong performance across models with varying structures and complexities.

\keywords{Time series classification\and Overfitting\and  Label smoothing}
\end{abstract}
\section{Introduction}
In recent years, time series tasks have gained increasing popularity, with applications spanning various fields including medical analysis~\cite{ref1}, stock market prediction~\cite{ref2}, weather forecasting~\cite{ref3}, and industrial scenarios~\cite{ref4}. A plethora of models have been developed and applied to such tasks. For instance, in the domain of time series classification, traditional machine learning algorithms like combing DTW with KNN~\cite{ref21} have shown promising results. And neural network (NN) has demonstrated its astonishing power in many data mining tasks~\cite{refadd1}.The advent of deep learning based models has further enriched the field, due to their superior performance and efficient computational speed. For example, Fawaz et al. introduced InceptionTime~\cite{ref5}, which has emerged as a powerful baseline model for time series classification tasks. LSTM-FCN~\cite{ref7}, developed by Fazle Karim et al., is notable for its end-to-end training capability and minimal preprocessing requirements. Neural network models leveraging residual connections, such as ResNet18~\cite{ref8}, have also been successfully applied in this context.

However, the overfitting phenomenon of deep learning based models in time series classification tasks has also attracted attention. Researchers have tried to analyze the reasons for overfitting from different angles. Chiyuan Zhang et al.~\cite{ref11}, Hassan Ismail Fawaz et al.~\cite{ref12} believe that the main reason is that the number of model parameters is too large. Xueyuan Gong et al.~\cite{ref14} conducted extensive experiments on the UCR datasets~\cite{ref39} to demonstrate the overfitting phenomenon, attributing this phenomenon mainly to the small sample size of the time series dataset, who used knowledge distillation~\cite{ref15} to soften the labels to alleviate the problem. Some model based modifications such as HIVE-COTE~\cite{ref16}, and ROCKET~\cite{ref17} can also alleviate this problem.

In addition to the opinions mentioned above, we analyze the reasons from two other aspects. Firstly, the soft label is more informative than the hard label because it can reflect the similarity between classes and the confidence of prediction~\cite{refHS}. For example, for a picture of a lion, in addition to the top 1 label should be a lion, a reasonable label should indicate that the picture has a higher degree of confidence in a relatively close animal like a tiger, rather than a shark, but hard labels will only treat other categories equally. Secondly, training the model with hard labels may lead to overconfidence in the model output distribution, thereby weakening the model's generalization ability. Christian Szegedy et al.~\cite{ref18} and Gabriel Pereyra et al.~\cite{ref19} have discussed the same problem in their work, and used label smoothing and confidence penalty to alleviate this problem respectively.

In this paper, we try to generate more reliable soft labels to improve the performance of models in the TSC task. Specifically, we use a time series encoder TS2Vec~\cite{ref20} pre-trained by contrastive learning~\cite{refcon} to build representations, and then construct soft labels based on the Euclidean distances. 
Our method can also be regarded as a label smoothing method, the implementation idea is similar to that of Online Label Smoothing~\cite{ref22}, both of which try to find soft labels that are more credible than using uniform distribution. But the latter constructs soft labels based on the output of the model itself, while ours based on the latent space of the encoder. Overall, the contributions of this article are as follows:

1. We propose a method that uses the encoder TS2Vec to generate sample representations combined with Euclidean distance to construct soft labels and perform label smoothing. 
Our method can be considered an improved form of label smoothing using an encoder, so it is not only applicable to the TSC task we studied but also holds potential for other tasks and domains, provided the encoder is adapted to the specific field.

2. We deployed our method on time series classification task, and introduced other 2 methods, label smoothing and confidence penalty, and baseline method using hard labels for comparison.

3. We trained 6 models with varying structures and complexities, using average accuracy as the evaluation metric. We also applied some visualization methods like critical difference diagrams and t-SNE method to compare the performance between different methods. The data results and visualization results demonstrate the improvement ability of our method from different perspectives.

\section{Related work}
\subsection{Time Series Classification}
We applied our method to the task of time series classification. Time series classification is a subfield of time series related tasks with numerous real-life applications~\cite{ref29}. Many methods have been applied to time series classification tasks and achieved competitive results. Patrick Schäfer et al.~\cite{refSVM} proposed a time series classification model based on SVM classifier. The KNN method combined with the DTW method is a very competitive method, but the time complexity of this method is high. Many researchers have proposed methods to accelerate DTW calculation of time series from different angles~\cite{ref30}~\cite{ref31}~\cite{ref32}. For deep learning based model, apart from the models like Inceptiontime, LSTM-FCN, Resnet18 that we mentioned before, there are also some other models like models based on attention mechanisms such as MACNN~\cite{ref9} and EMAN~\cite{ref10}.
\subsection{Hard Label and Label-Related Regularization Methods}
For a classification task with T categories, the hard label is typically represented as an one-hot encoding vector, where only the category corresponding to the ground truth is 1, and all other categories are 0, like $y_i = [0, 0, \ldots, 1, \ldots, 0]^T$. Szegedy et al. pointed out that such labels may cause the trained model to be overly confident, leading to overfitting and a reduction in generalization ability. They proposed label smoothing to obtain a less confident soft label for training. Pereyra et al. proposed the confidence penalty which aims to punish a low-entropy model by subtracting a weighted entropy term from the loss function, encouraging the output distribution of the model to be more dispersed. There are also some other regularization techniques,  Bootstrapping~\cite{ref23} achieves regularization by using the predicted distribution and class. Xie et al.~\cite{ref24} implements regularization by randomly replacing a portion of labels with incorrect values. Our method is similar to the idea of label smoothing, which weakens the confidence of hard labels to prevent the model from being overly confident. 

\subsection{Knowledge Distillation}
Knowledge distillation is a popular technique for model compression. It significantly improves the performance of the student model by letting it learn soft labels provided by the pre-trained teacher model. This technology has been widely used in specific application scenarios such as language model~\cite{ref25} and federated learning~\cite{ref26}. We use a similar loss function structure to knowledge distillation, but our method falls under label smoothing, which can be widely used to improve model generalization capabilities and reduce overfitting, while knowledge distillation is generally employed when training lightweight models. And typical knowledge distillation requires the teacher model and the student model to have similar structures, while we do not need the encoder to be structurally similar to the model involved in training, which means our method is more flexible.

\section{Methodology}
\subsection{Preliminaries}
\subsubsection{Label smoothing}
Label smoothing processes the distribution of hard labels to change the confidence of sample labels. For a training example with ground-truth label $y$, the label distribution $q(k|x) = \delta_{k,y}$ are replaced with 
$q'(k|x) = (1 - \varepsilon)\delta_{k,y} + \varepsilon u(k)$, 
which is a mixture of the original ground-truth distribution $q(k|x)$ and the uniform distribution $u(k)$, with weights $1 – \varepsilon$ and $\varepsilon$. 
For model $P_{\theta}$, minimizing the cross-entropy loss $L_{\text{ls}}(\theta)$ of such a modified distribution is the same as adding a weighted KL divergence between uniform distribution $u$ and the model output $P_{\theta}$. Therefore, the loss function could be rewritten as~\cite{ref37}:
\[ L_{\text{ls}}(\theta) = L(\theta) + \beta D_{\text{KL}}(u||P_{\theta})\]
\subsubsection{Confidence penalty}
Confidence penalty regularizes the model by penalizing the confidence level of the output distribution, thereby preventing the model from being too confident. Pereyra et al. systematically explored the regularization effect of using confidence penalty, who added a weighted entropy term to the model's loss function, which can reflect the concentration of the model's output distribution. With this approach, they prevent the distribution generated by the model from being too concentrated. This method is equivalent to adding a KL divergence to measure the difference between the model output and the uniform distribution. The loss function could be expressed as:
\[ L_{\text{cp}}(\theta) = L(\theta) + \beta D_{\text{KL}}(P_{\theta}||u) \]

\subsubsection{Knowledge distillation}
In knowledge distillation, the soft labels of the teacher model will be used as additional information to train the student model. In this process, KL divergence is used to align the softmax output of the teacher model and the student model. For a multi-classification task with $L$ categories, assume that $F_s(x)$ represents the student model and $F_t(x)$ represents the teacher model. They take $x$ as input and output an $L$-dimensional vector $P$. Then $F_t(x) = P_t = \text{softmax}(a_t)$, $F_s(x) = P_s = \text{softmax}(a_s)$, where $a$ is the pre-softmax activation. In the student model training, the temperature coefficient $\tau$ is introduced to soften the output of softmax function, and the coefficient \(\lambda\) determines the weight of the KL divergence. The loss function can be written as:
\[ L = L_{\text{CE}}(P_{s}, y) + \beta D_{\text{KL}}(P^{\tau}_{t}, P^{\tau}_{s}) \]

\subsection{Proposed Method}
\begin{figure}[!h]
    \centering
    \includegraphics[scale=0.6]{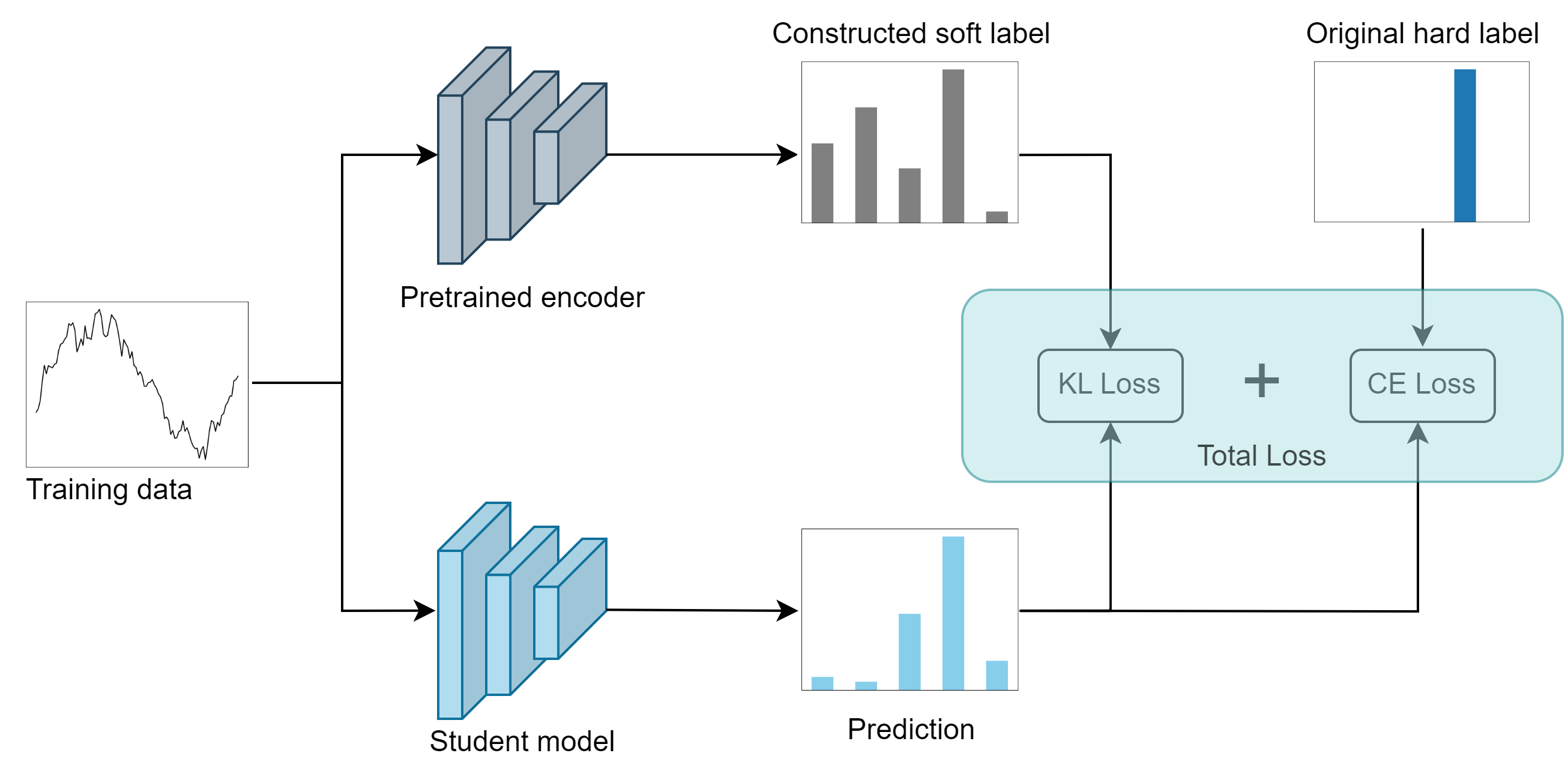}
    \caption{The structure of our method.}
    \label{3}
\end{figure}

In this section, We introduce our method of constructing soft labels and performing label smoothing. The structure of our method could be found in Figure.~\ref{3}. Firstly we considered using DTW method to directly construct soft labels, but as we mentioned before, the time complexity of this method is too high. Directly using L2 distance is another option and much faster, but when comparing two time series with similar patterns but different phases, using L2 distance directly might result in a large error. While TS2Vec trained an encoder using unsupervised contrastive learning to generate sample representations for downstream tasks and achieved good results, we use TS2Vec to obtain the representation $r_i$ of the sample, and use the L2 distance $\|r_i - r_j\|$ of the representation of the sample and samples of different categories as a measure of the confidence to generate soft labels. 

\begin{figure}[!h]
    \centering
    \includegraphics[scale=0.45]{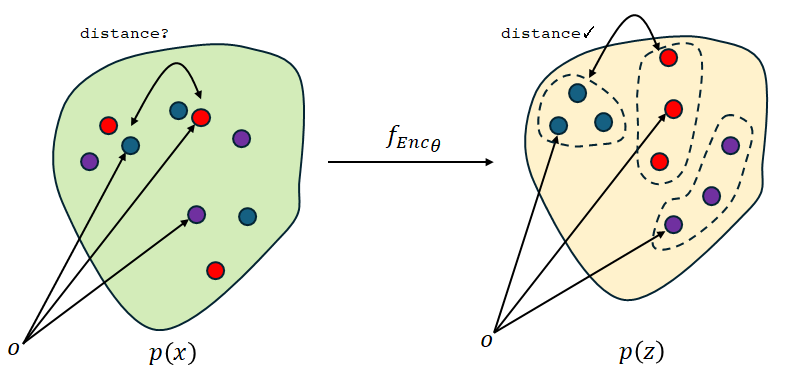}
    \caption{The use of an encoder generates different representation spaces for the samples, which are measured using L2 distance later.}
    \label{fig3a}
\end{figure}

\begin{figure}[!h]
    \centering
    \includegraphics[scale=0.45]{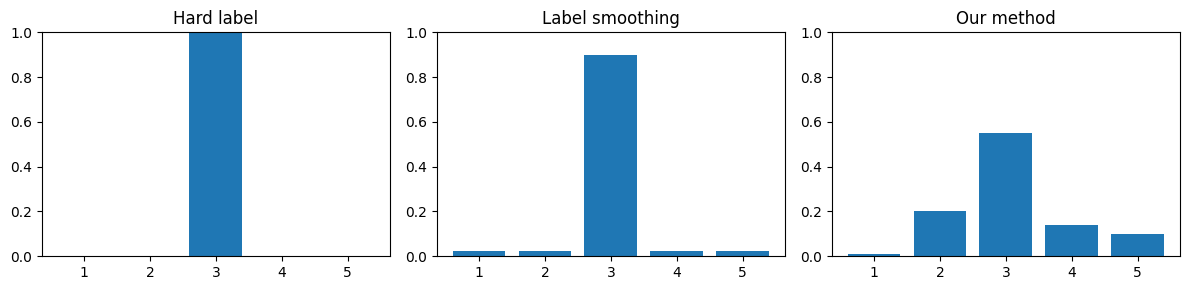}
    \caption{Hard labels, labels produced by label smoothing and labels produced by our method.}
    \label{fig3b}
\end{figure}

A similar structure to knowledge distillation is then used to allow the student model to receive soft labels and original data for training. The soft labels constructed using this method contain the distance information between different classes of samples in the encoder's latent space, which are regarded as the similarity information. Fig.~\ref{fig3a} illustrates the difference in using the encoder to process data before and after, with the use of L2 distance for measurement. Fig.~\ref{fig3b} shows different soft labels produced by different methods. Our loss function can be expressed as:
\[ L = L_{\text{CE}}(P_{s}, y) + \beta D_{\text{KL}}(P^{\tau}_{E}, P^{\tau}_{s}) \]
where $P^{\tau}_{E}$ represents the soft labels constructed by the encoder and are balanced by temperature coefficient.

\subsubsection{Sample confidence and soft label construction} 
The inspiration for using L2 distance to measure the similarity between samples and different classes comes from Z.Wang~\cite{ref38}. The author originally aimed to quantify sample robustness in computer vision classification tasks by using the L2 minimum distance between samples and samples from other classes.We made the following improvements based on the original method. Initially, when presenting the author's approach for assessing robustness between samples and classes, it was noted that directly calculating the L2 distance between time series samples might result in significant errors due to phase issues like we mentioned before. So we use the L2 distance between samples calculated from the representations processed by the encoder. The author directly used the minimum distance from one sample to samples of other classes as a measure of sample robustness, which is easily affected by individual outlier samples. Instead, we use the average L2 norm of the sample $m_{i}$ to samples of other classes n as a measure of the confidence. The formula is expressed as $||m_{i},n||= (||m_{i}-n_{1}||+||m_{i}-n_{2}||+…+||m_{i}-n_{n}||)/n$. The larger the average distance is, the larger the distance between the sample and the target class is in the latent space, and the less similar the sample is to this class. After obtaining the average distance between all samples of different classes, the soft labels are generated based on the following two criteria.

\begin{theorem}
Assuming the correct category of sample $m_i$ is A, then $\text{argmax}\, (m_i) = A$. We must ensure that the correct category in the constructed soft label has the highest confidence, so that the sample is classified into the correct category.
\end{theorem}

\begin{theorem}
The larger the average distance $r_{m,n}$ between a sample and other classes, the smaller the confidence of the class ${a}_{m,n}$ in the soft label of the sample should be. This is because the larger the distance $r_{m,n}$, the less likely the sample belongs to this class.
\end{theorem}

Following these two rules, we design the following soft label generation method. Here $\gamma$ is a uniform coefficient designed according to the characteristics of the time series samples.

\[
{a}_{m,n} =
\begin{cases}
\gamma\times \frac{1}{r_{m,n}}, & \text{for } n \neq m, \\
\sum\limits_{i=1}^{L} \gamma\times\frac{1}{r_{m,i}}, & i \neq m, \text{ for } n = m.
\end{cases}
\]

A location's confidence, not belonging to the correct class, is inversely proportional to the average distance of a sample to samples of that class. In order to ensure that the item with the highest soft label confidence is the correct category, we set its confidence to the sum of the confidences of other classes. Finally, for sample $m_{i}$, we get its soft label through $m_{i}=softmax(a_{m}$). For our algorithm flow, see Algorithm 1. 

\begin{algorithm}
\caption{Representation Soft Label Smoothing}
\label{alg:knowledge_distillation}
\textbf{Input:} Training dataset \(D_{train}\). Pre-trained Encoder \(E\). Untrained Student model \(S\).

\textbf{Output:} Trained Student model \(S\).

\textbf{Hyperparameters:} Distillation temperature \(\tau\). Weight of KL divergence \(\beta\).

\textbf{Method:}

\textbf{Step 1}
\begin{algorithmic}[1]
\FOR{\(sample\) in \(D_{train}\)}
    \STATE Get the representation of the sample: \(sample' = E(sample)\).
\ENDFOR

\FOR{each \(sample'_{ix}\) in class \(i\)}
    \FOR{each \(sample'_{jx}\) in class \(j\)}
        \STATE Compute \(||sample'_{ix}, sample'_{jx}||\).
    \ENDFOR
    \STATE Compute \(||sample'_{ix}, j||\) as the average \(L2\) distance of \(sample'_{ix}\) to samples in class \(j\).
    \STATE Set \(a_{ix,j} =  1 /||sample'_{ix}, j||\times\gamma\).$(j \neq i)$
    \STATE Set \(a_{ix,i} = \sum_{j \neq i} a_{ix,j}\).  Then we get the complete distribution $a_{ix}$.
    \STATE Get output $m_{ix} = \text{softmax}(a_{ix})$.
\ENDFOR
\end{algorithmic}

\textbf{Step 2}
\begin{algorithmic}[1]
\STATE Initialize Student model \(S\). 
\FOR{\(epoch = 1\) to \(1000\)}
    \FORALL{\(x, y\) in \(D_{train}\)}
        \STATE Introduce and soften the corresponding soft label: \(P^{\tau}_{E} = m_{ix} / \tau\).
        \STATE Compute Student logits: \(P_S = S(x)\).
        \STATE Compute loss: \( L = L_{\text{CE}}(P_{s}, y) + \beta D_{\text{KL}}(P^{\tau}_{E}, P^{\tau}_{s}) \)
        \STATE Update Student model: \(S = S - \eta \cdot \nabla L\).
    \ENDFOR
\ENDFOR
\RETURN \(S\)
\end{algorithmic}
\end{algorithm}

\section{Experiment Setup}
\subsection{Experimental Environment}
The experiments were conducted on a computer equipped with an intel core i7 13700K, 64GB of memory and an NVIDIA RTX 2080ti GPU. 

\subsection{Dataset}
The UCRArchive2018 was used during the experiment. This dataset collects 128 time series related sub-datasets from different fields, different lengths and different numbers of categories, and has become a well-known benchmark in time series classification. The samples in the dataset have been divided into two parts: training set and test set.

\subsection{Model Selection}
 A total of 6 models with different structures and different complexities were selected for experiment. For different network structures, 3 models were selected, including Inceptiontime, LSTM-FCN and Resnet18, which are commonly used in time series classification tasks. For models of different complexity, Inceptiontime-3, Inceptiontime-2 and Inceptiontime-1 which contain 3, 2 and 1 inception modules respectively, were selected. Accuracy was chosen as the metric to evaluate the performance of classification models.

\subsection{Hyperparameter Setting}
Different methods utilize various parameter combinations. For label smoothing, the parameter $\beta$ determines the smoothing strength of the label. We adopt the original parameter $\beta$=0.1 used by the author. For the confidence penalty, the parameter $\beta$ determines the weight of the confidence penalty term. The coefficient we chose is 0.1, which is one of the parameters that the author achieved the best result in the classification tasks.

In our method, the parameter composition is the same as that of knowledge distillation, the temperature coefficient $\tau$ and the weight $\beta$ of the KL divergence term. For different models, we experimented with temperature coefficient combinations [2, 4, 10] and KL divergence weight combinations [0.1, 0.5, 1] and selected parameters from them. For the temperature coefficient term, except Resnet18 which is 4, the temperature coefficient of all other models is $\tau$=2. For the weight $\beta$ of KL divergence, Inceptiontime takes 1, Inceptiontime-1 and Inceptiontime-2 take 0.5, and Inceptiontime-3, LSTM-FCN, and Resnet18 take 0.1.

During training, we typically trained all models for 1000 epochs. We performed forward propagation on the test set every 5 training epochs, recording the best performance for comparison. Due to equipment limitations and the principle that larger batch sizes usually lead to better model performance, we uniformly set the batch size to 128. We used the Adam optimizer with an initial learning rate of 0.001. For the uniform coefficient $\gamma$, we set $\gamma$=0.001 in all experiments.

\section{Experimental Results}
In this section, we compared the performance of 6 models under baseline, label smoothing, confidence penalty and representation soft label smoothing. We also conducted ablation study on whether to perform label smoothing and whether to use the encoder to process samples. In the experimental results, label smoothing is abbreviated as LS, confidence penalty is abbreviated as CP, and our method representation soft label smoothing is abbreviated as SS. 
\subsection{Performance Evaluation}
We compared the average accuracy of models across 4 methods on UCR datasets. For the Inceptiontime model, our method achieved the highest average accuracy of 0.8555, which was 0.0068 higher than the baseline. The CP method had the second-highest accuracy, followed by the baseline, and then LS.
For LSTM-FCN, our method also outperformed the others with an average accuracy of 0.827, which was 0.015 higher than the baseline. The CP method followed, then LS, and finally the baseline.
For Resnet18, our method had the highest average accuracy of 0.8195, 0.0144 higher than the baseline. LS was the second-highest, followed by CP, and then the baseline.
Inceptiontime-3 showed that our method had the highest average accuracy of 0.8483, which was 0.0045 higher than the baseline. LS was the second-highest, followed by the baseline, and then CP.
Inceptiontime-2 displayed our method with the highest average accuracy of 0.835, which was 0.0348 higher than the baseline. LS followed, then CP, and finally the baseline. It's notable that our method demonstrated a significant improvement in accuracy, exceeding 0.02 compared to the other 3 methods.
Lastly, for Inceptiontime-1, our method showcased the highest average accuracy of 0.7318, which was 0.0714 higher than the baseline. The CP method came next, followed by LS, and then the baseline. Our method demonstrated the most substantial improvement in accuracy among the models, with an increase of over 0.06 compared to the other 3 methods.

From the experimental results we can see in most cases, using methods that weaken the confidence of the model output can produce better results than directly using soft labels.The comparison of the average accuracy of different methods under the different models is summarized in Table~\ref{table1}.

\begin{table}
\centering
\caption{Comparison of average accuracy of different situations}\label{table1}
\setlength{\tabcolsep}{7pt} %
\renewcommand{\arraystretch}{1} %
\resizebox{0.8\textwidth}{!}{%
\begin{tabular}{c|c|c|c|c}
\hline
model/method & baseline & SS(Ours) & LS & CP\\
\hline
Resnet18 & 0.8051 & \textbf{0.8195} & 0.8111 & 0.8052\\ 
LSTM-FCN & 0.8120 & \textbf{0.8270} & 0.8140 & 0.8170\\ 
Inceptiontime & 0.8487 & \textbf{0.8555} & 0.8462 & 0.8497\\ 
Inceptiontime-3 & 0.8438 & \textbf{0.8483} & 0.8455 & 0.8432\\ 
Inceptiontime-2 & 0.8002 & \textbf{0.8350} & 0.8054 & 0.8045\\ 
Inceptiontime-1 & 0.6604 & \textbf{0.7318} & 0.6638 & 0.6661\\ 
\hline
\end{tabular}
}
\end{table}
 
\begin{figure}
    \centering
    \includegraphics[scale=0.21, trim=50 0 0 0]{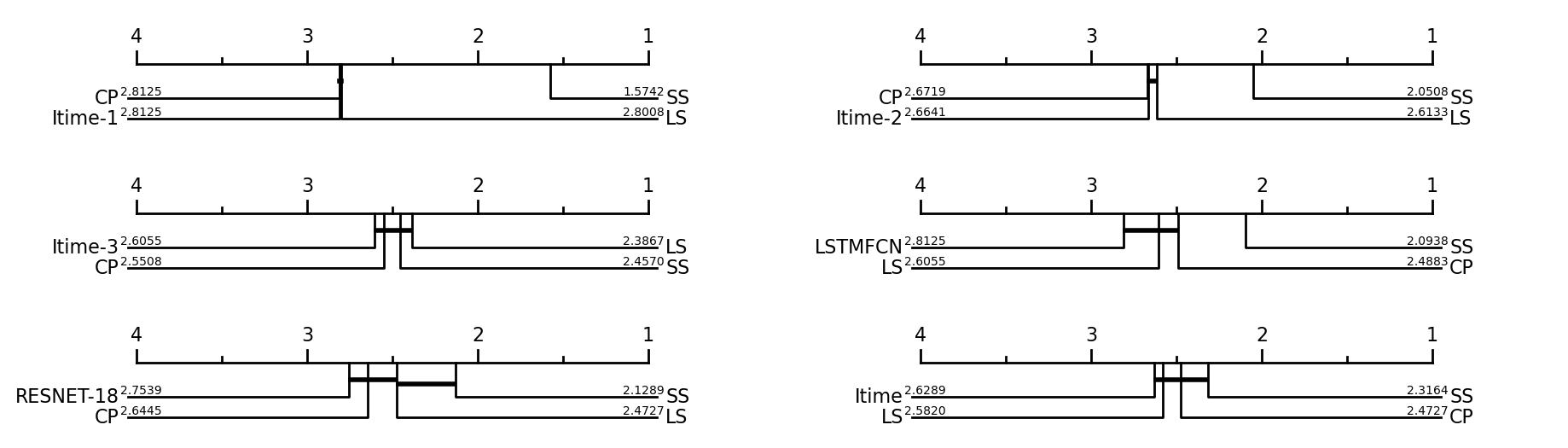}
    \caption{Critical difference diagrams for 6 models with different methods} \label{CD1}
\end{figure}

We also utilized critical difference diagrams to conduct ranking analysis on the data from the 4 methods across different models. The results are depicted in Fig.~\ref{CD1}. It is evident that in 5 out of the 6 groups of models, the rankings of CP, LS, and our method are better than or equal to the ranking of the baseline. This indicates that in most cases, methods that reduce model confidence outperform models trained directly with hard labels. Additionally, our approach achieves the best ranking in most cases. Only in the case of the Inceptiontime-3 model is it surpassed by the LS method and ranked second. This highlights the competitiveness of our method among the improved methods.

\begin{figure}[!h]
\centering
\includegraphics[scale=0.2]{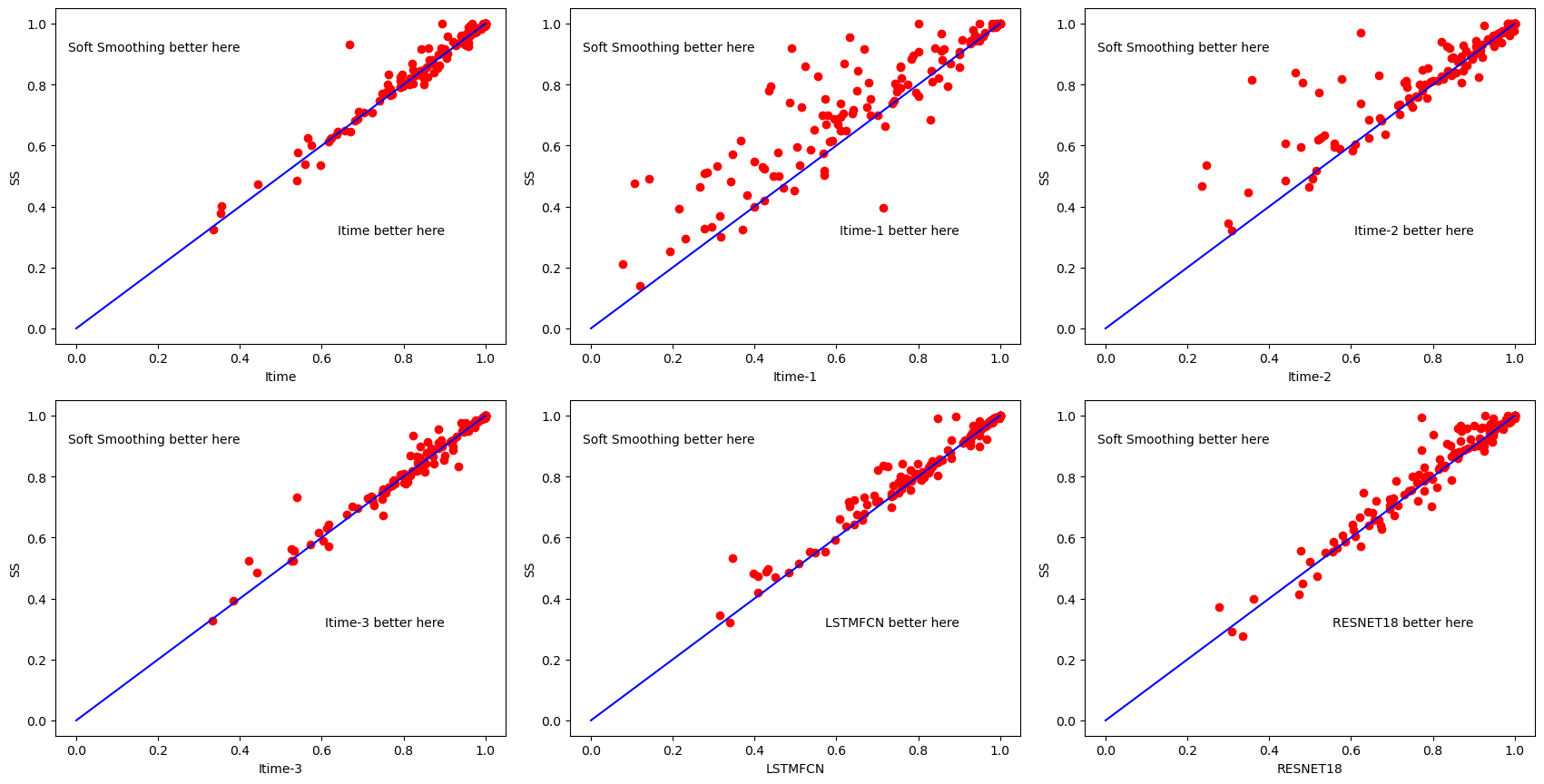}
\caption{Comparison of accuracy for 128 sub-datasets between our method and baselines}
\label{com}
\end{figure}

We aimed to compare the outcomes of employing our method versus directly training a model with hard labels. To do so, we created visuals illustrating the accuracy of each sub-dataset for 6 models in detail. As depicted in Fig.~\ref{com}, our method's models generally outperform those models trained with hard labels, particularly evident in Inceptiontime-1 and Inceptiontime-2 models. This observation underscores the competitiveness of our approach, particularly when applied to lower-complexity models, when contrasted with the baseline models.

\subsection{Visualization Comparison} 
\begin{figure}[!h]
\centering
\includegraphics[scale=0.15]{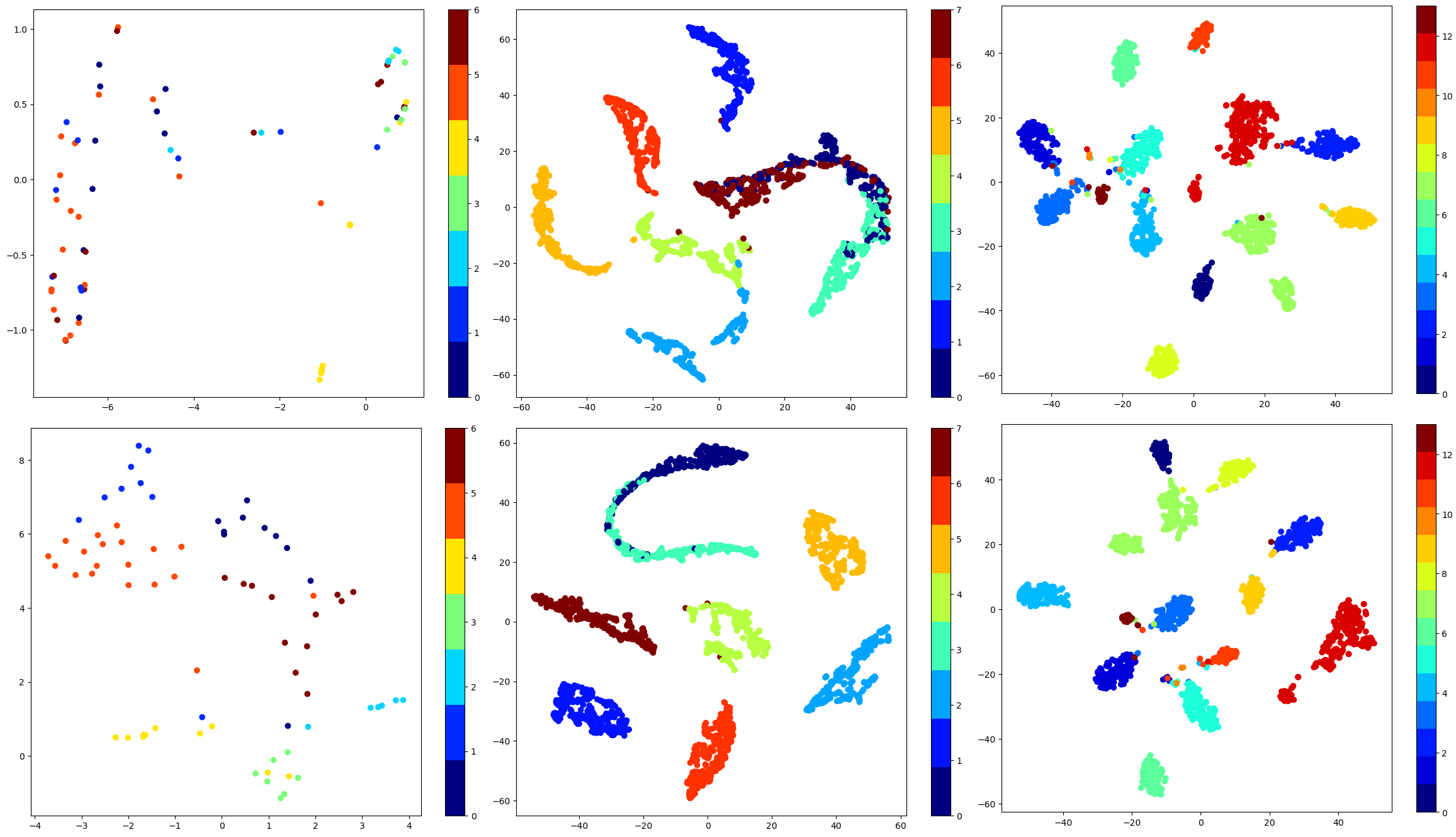}
\caption{t-SNE visualization.The top side shows the baseline result, while the bottom side shows our method. From left to right, they are Inceptiontime-1 under Lightning7 dataset, Inceptiontime-2 under Mallat dataset, and Resnet18 under FaceAll dataset.}
\label{fig:cd1}
\end{figure}

In Fig.~\ref{fig:cd1}, we utilized the t-SNE method to reduce the dimensionality of features to 2, allowing us to visualize some of the datasets used in models with both the baseline method and our approach. With 128 datasets containing various test samples and numbers of classes in the UCR, visualizing all datasets is impractical. Experimental results demonstrate that our proposed approach improves the distinguishability between representations of different classes. Two key observations support this conclusion. Firstly, in the comparison of the first column of images, it is evident that the data points of the same category in the t-SNE diagram generated by our method are more tightly clustered compared to those using the baseline method. Secondly, when comparing the second and third columns, the spacing between data points of different categories is more distinct in our method's visualization.

\subsection{Ablation Study}
In order to verify the effectiveness of different modules, we conducted ablation experiments on 6 models. The experiments were divided into 3 groups: The first group trained the model using our method, the second group did not use an encoder and directly used the original samples combined with L2 norm to construct soft labels for label smoothing, and the third group did not use label smoothing and only trained the model with hard labels. Table 2 shows the results of the ablation experiment. It can be seen that in all 6 models, our method achieves the best results, better than the cases without encoder and without label smoothing. This shows that using encoder to construct soft labels and perform label smoothing is the optimal combination.
\begin{table}
\centering
\caption{Comparison of average accuracy of different situations in ablation study}\label{table2}
\setlength{\tabcolsep}{7pt} %
\renewcommand{\arraystretch}{1} %
\resizebox{0.8\textwidth}{!}{%
\begin{tabular}{c|c|c|c}
\hline
model/method & SS(Ours) & w/o encoder & w/o soft label \\
\hline
Resnet18 & \textbf{0.8195} & 0.8056 & 0.8051 \\ 
LSTM-FCN & \textbf{0.8270} & 0.8242 & 0.8120 \\ 
Inceptiontime & \textbf{0.8555} & 0.8473 & 0.8487 \\ 
Inceptiontime-3 & \textbf{0.8483} & 0.8464 & 0.8438 \\ 
Inceptiontime-2 & \textbf{0.8350} & 0.8308 & 0.8002 \\ 
Inceptiontime-1 & \textbf{0.7318} & 0.7305 & 0.6604 \\ 
\hline
\end{tabular}
}
\end{table}

\section{Conclusion}
In this paper, we propose a new label smoothing method, which is called representation soft label smoothing. This method can be viewed as an improved version of traditional label smoothing, which aims to create more reliable soft labels and is used to reduce the confidence of the model's output to achieve better results. At the same time, some regularization methods like label smoothing and confidence penalty are introduced into time series classification, and we compare them with the baseline together with our method. 

The experiments demonstrate the effectiveness of methods like label smoothing in reducing model confidence in time series classification tasks, showing competitive results compared to the baseline. Representation soft label smoothing achieves the best average accuracy and ranking in most cases, highlighting its superiority. Specifically, experiments with the simpler models, Inceptiontime-1 and Inceptiontime-2, showcase the potential of our method to enhance model performance in simpler model structures.

In the future work, we will continue to experiment the soft labels generated by encoders with different structures, and integrate them with various models and fields for downstream tasks to explore the versatility of our method. Additionally, we will further investigate the effects of our method on improving model performance across different levels of complexity.

\section{Acknowledgements}
We would like to express our gratitude to Chang Dong,Yi Han, Zhengyang Li and Liangwei Zheng (The University of Adelaide, Australia) for their invaluable support on the experimental approach throughout this work.

\end{document}